\def\year{2020}\relax
\documentclass[letterpaper]{article} 
\usepackage{aaai20}  
\usepackage{times}
\usepackage{latexsym}
\usepackage{algorithm}
\usepackage{algorithmic}
\usepackage{amsmath}
\usepackage{latexsym}
\usepackage{tablefootnote}
\usepackage{pdflscape}
\usepackage{booktabs}
\usepackage{soul}
\usepackage{xcolor,colortbl}
\usepackage{amsfonts}
\usepackage[english]{babel}
\usepackage{tikz-dependency}
\usepackage{tikz}
\usepackage{tikz-qtree}
\usepackage{tablefootnote}
\usepackage{multibib}
\usepackage{algorithm}
\usepackage{wrapfig,lipsum,booktabs,multirow,subcaption,multicol}
\usepackage{xcolor,colortbl,framed,cuted,tcolorbox}
\usepackage{pbox}
\usepackage{times}  
\usepackage{helvet} 
\usepackage{courier}  
\usepackage[hyphens]{url}  
\usepackage{graphicx} 
\usepackage{graphicx}  
\title{MetaMT, a Meta Learning Method Leveraging Multiple Domain Data for Low Resource Machine Translation }
\author{Rumeng Li\textsuperscript{\rm 1} \thanks{Equal contribution}, Xun Wang\textsuperscript{\rm 1,2,3} \footnotemark[1], Hong Yu\textsuperscript{\rm 1,2,3,4}\\
\textsuperscript{\rm 1}School of Computer Science, University of Massachusetts Amherst, Amherst, MA, United States\\
\textsuperscript{\rm 2}Department of Computer Science, University of Massachusetts Lowell, Lowell, MA, United States\\
\textsuperscript{\rm 3}Bedford Veterans Affairs Medical Center, Bedford, MA, United States\\
\textsuperscript{\rm 4}Department of Medicine, University of Massachusetts Medical School, Worcester, MA, United States\\
alicerumeng@foxmail.com,wangxun@outlook.com,Hong\_Yu@uml.edu 
}

\begin{document}

\maketitle

\begin{abstract}
Neural machine translation (NMT) models have achieved state-of-the-art translation quality with a large quantity of parallel corpora available. However, their performance suffers significantly when it comes to domain-specific translations, in which training data are usually scarce.
In this paper, we present a novel NMT model with a new word embedding transition technique for fast domain adaption. We propose to split parameters in the model into two groups: model parameters and meta parameters. The former are used to model the translation while the latter are used to adjust the representational space to generalize the model to different domains. We mimic the domain adaptation of the machine translation model to low-resource domains using multiple translation tasks on different domains. A new training strategy based on meta-learning is developed along with the proposed model to update the model parameters and meta parameters alternately. 
Experiments on datasets of different domains showed substantial improvements of NMT performances on a limited amount of data.
\end{abstract}

\section{Introduction}
Neural machine translation (NMT) is a deep learning-based approach for machine translation. A typical NMT model comprises three parts: the encoder, the decoder and the attention model. It relies on large quantities of parallel corpora for effective training of the unprecedented amount of parameters in each of its parts and to avoid overfitting \cite{bahdanau2014neural,gehring2017convolutional,vaswani2017attention,meng2019large}. However, finding such data remains challenging for specific domains as the construction of parallel corpus is often too expensive. The scarcity of domain-specific parallel data limits the potentials of NMT models in low-resource scenarios as previous works stated \cite{zoph2016transfer,gu2018meta,koehn2017six}.

One research direction for resource lean approaches is to leverage data of multiple domains to develop robust translation systems that can be migrated to specific domains easily.

When migrating NMT models from one domain to another, one of the biggest challenges faced by researchers  is the domain divergence.
Domain divergence causes difficulties for NMT in at least two aspects. Firstly, different domains tend to have their own distinct set of vocabulary. For example, when referring to the same pathology, a medical corpus would be inclined to use ``cardiovascular disease", while ``heart disease" would be more common among corpora of general domains. The resulting large vocabulary aggravates the problem of data sparsity. Secondly, even identical words may carry different meanings in the context of their respective domains. For example, ``Obama" is widely known as the former president of US, but the same word usually refers to a beautiful seaside city in Japan in a corpus about Japan. As such, embeddings for the same word cannot be shared across domains. This is referred to as the polysemy problem which has been addressed by many works \cite{yarowsky1996three,chen1999resolving,koehn2009statistical} and this problem is more severe for domain-specific machine translation. 

Existing approaches for domain adaptation NMT generally fall into two categories: data-centric and model-centric \cite{chu2018survey}.  Data-centric methods focus on creating more data from either in-domain monolingual corpora, synthetic corpora or parallel corpora \cite{zhang2016exploiting,domhan2017using,chu2017empirical}. The creation of more in-domain training data could balance the ratio between the in- and out-of-domain data and therefore enable the learnt model to pay more attention to the target domain.
The model-centric category focuses on NMT models that are specialized for domain adaptation such as Fine Tuning \cite{dakwale2017fine} and Instance/Cost Weighting \cite{wang2017instance,D17-1155}. Such methods could also leverage out-of-domain parallel corpora or in-domain monolingual corpora. Fine tuning places data of the target domain at the end of the training data stream, forcing the model to pay more attention to the target domain. Instance/Cost Weighting methods force the model to focus on the target domain by explicitly assigning higher weights to data in the target domain (or similar  to the data in the target domain) during training.
Both instance/cost weighting and fine tuning optimize the model towards a local minimum which benefits target-specific performances \cite{DBLP:journals/corr/KocmiB17aa}. Previous work shows the above methods yield similar improvements \cite{wang2017instance}.

Inspired by existing work, we propose to manipulate the training data of multiple domains to mimic domain adaptation and train a novel model which addresses the big vocabulary and polysemy problem. Instead of using one large lookup table to store all word representations, the designed model firstly projects all words to a semantic space that is shared by all domains. In this shared semantic space, one word is represented by a selected number of base words. This helps reduce the vocabulary size and also enables words of different domains to have different representations. A transmission layer is used in our model to conduct the mapping of word vectors.

We repeatedly train the model using a (relatively) large dataset of one domain and fine tune it on another domain with a small dataset. We adopt a meta learning strategy which enables fast parameter adaptation on small datasets. Two kinds of parameters are defined in our model as meta parameters and model parameters.
The model parameters are used to learn the translation from the source sentences to the target sentences. The meta parameters are used to enhance the generalization ability of the learnt model. At fine tuning, we freeze the model parameters and adjust only the meta parameters. The meta learning strategy \cite{finn2017model} acts as to learn a parameter initialization that can be quickly adopted to new domains.

As there are no language specific features required  in the proposed method, it can be applied to any language pairs. For our evaluation we focus on the translation of two mostly widely spoken language pairs as English to Spanish translation, for which we both have a handful of datasets of different domains which can be accessed easily. Experiments show that the proposed method improved results when evaluated using BLEU \cite{papineni2002bleu} as compared to existing transfer-learning NMT methods. To further verify the effectiveness of the proposed model, we use a small dataset with only three thousands sentences of electronic health records. Experiments show that the proposed model can produce high quality results for the specific domain when trained on thousands of sentences.

The contribution of this work is two-fold: firstly, a novel domain adaption training strategy based on the meta-learning policy is proposed for neural machine translation. Secondly, a novel word embedding transition technique is proposed to help handle domain divergence.

\section{Background}
\subsection{The Encoder-Decoder Model and NMT}
The encoder-decoder model has been used as the backbone for a wide range of NLP generation tasks including machine translation \cite{bahdanau2014neural,vaswani2017attention}, summarization \cite{rush2015neural,nallapati2016abstractive}
and dialogue generation,
\cite{li2017adversarial,baheti2018generating}.

The encoder-decoder model transforms a source sentence $s=(w^s_1,w^s_2,...,w^s_m)$ into a target sentence $t=(w^t_1,w^t_2,...,w^t_n)$ with neural networks as follows.
\begin{equation}
P(t|s)=\prod_{i=1}^{n}p(w^t_i|w^t_{<i},s)
\end{equation}
 
The conditional probability of $P(t|s)$ is parameterized by the encoder-decoder framework. The encoder generates vector representations from a variable-length input sentence, and the decoder outputs a correct translation correspondingly using these vector representations.

Typical structures employed for the encoder-decoder architecture include RNN \cite{schuster1997bidirectional,mikolov2010recurrent}, 
recursive tree structures \cite{liu2014recursive,li2015tree},  
LSTM \cite{hochreiter1997long} and GRU \cite{cho2014properties} for better handling of long-dependency.
A significant characteristic of recent NMT models is the wide use of attention mechanism \cite{bahdanau2014neural,li2016simple,vaswani2017attention}. Attention mechanism is firstly used in the decoder to enable the decoder to look at the input again and choose the most relevant parts to attend to at each step in translation. Later works like the transformer model \cite{vaswani2017attention} extensively employ the attention mechanism at both the encoder and the decoder sides to capture semantic relations inside sentences. Our proposed model is based on the transformer model, and the detailed description of which can be found in \cite{vaswani2017attention}.

\subsection{Domain adaptation for NMT}
Data sparsity has long been a problem for NMT. Domain adaptation methods are employed for NMT when the amount of in-domain parallel corpora is insufficient for training a good NMT system. Conventional ways for domain adaptation are fine-tuning \cite{luong2015stanford,sennrich2016edinburgh} where models are first trained on a high-resource domain or a mixture of data of different domains to initialize parameters which is further trained on the low-resource domain. In-domain fine-tuning comes with at least two shortcomings: firstly, it depends on the availability of sufficient amounts of in-domain data to avoid over-fitting; secondly, it results in degraded performance for all other domains.
Curriculum learning has also been exploited \cite{zhang2019curriculum} for domain adaptation. As proved in previous work \cite{bengio2009curriculum}, adjusting the order of training data leads to improvements in both the convergence speed and performances.
\cite{wuebker2018compact} studied the fine tuning process and pointed out that it is possible to do domain adaptation by tuning only a small proportion of the model parameters. This strategy has been adopted by our work by splitting parameters into meta parameters and model parameters.
\cite{zeng2018multi} proposed to generate domain-specific and domain-general representations for words. \cite{vilar2018learning} proposed that different neurons play different roles in different domains. It is thus necessary to adjust neurons weights according to data. Instead of manipulating neurons or word representations, we use a neural mapping to consider domain divergence.

In this paper we work on the translation of English to Spanish language in different domains. A novel learning policy based on meta learning is proposed to work with the designed model. Details are explained in the following section. 

\subsection{Meta Learning}
Meta learning has been drawing much attention of the NLP research community recently due to its ability in learning to transfer knowledge across tasks and domains \cite{finn2017model,hochreiter2001learning}. 

Meta learning, also known as ``learn to learn", intends to make machine learning models adaptive to a broader category of tasks/datasets other than the ones they are designed for or trained on. 
From the perspective of meta learning, training can be regarded as learning a prior over model parameters that is capable for fast adaptation on a new task/dataset. 

Current meta learning methods in machine learning refer to a broad category of learning strategies and policies. Works on this topic can be roughly grouped into the following two kinds:
1) Meta learning as a policy, such as transfer learning \cite{sennrich2016edinburgh,gu2018meta,finn2017model} and learning curriculum\cite{DBLP:journals/corr/KocmiB17aa}.
2) Meta learning as a parameter updating algorithm \cite{hochreiter2001learning,DBLP:journals/corr/MunkhdalaiY17,andrychowicz2016learning}.
\begin{figure}[ht]
  \centering
  \includegraphics[width=0.49\textwidth]{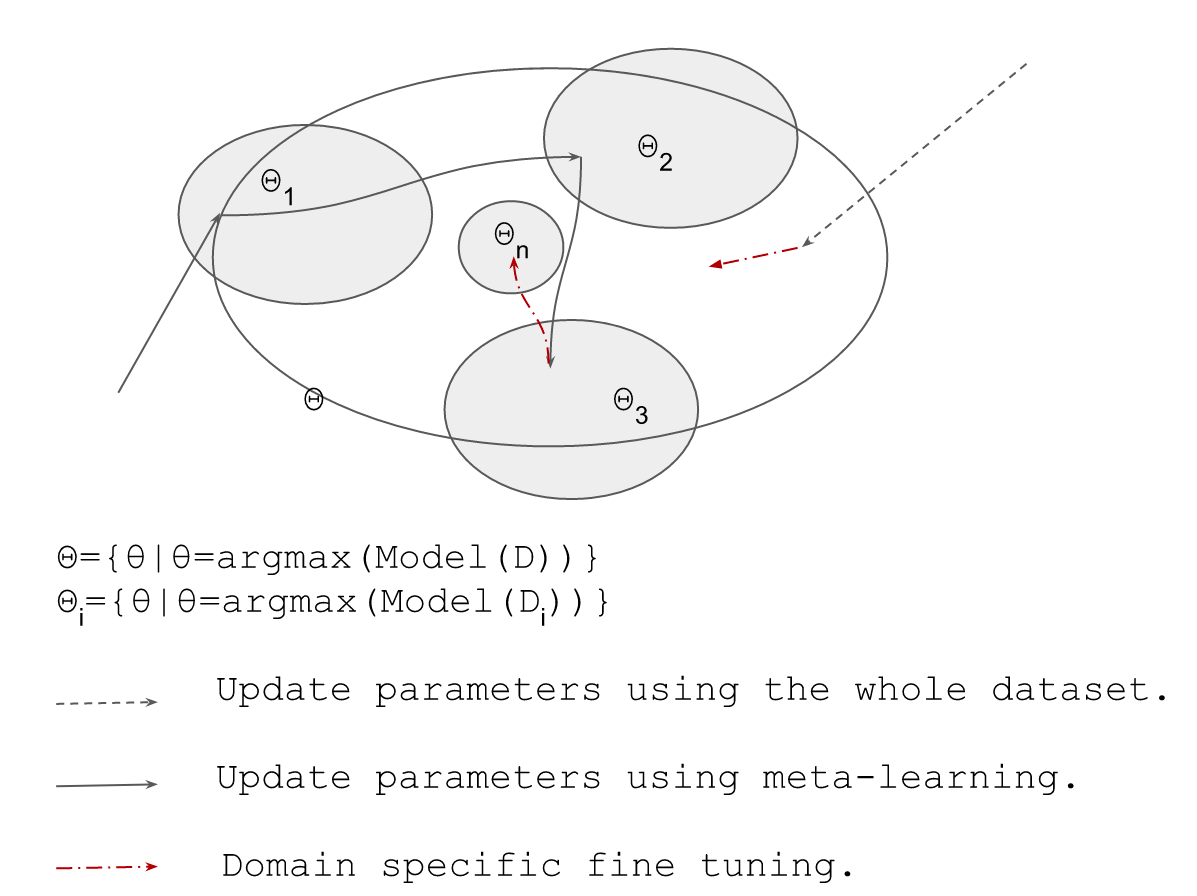}
  \caption{A graphical illustration of the parameter updating procedures in meta learning and fine tuning.}\label{fig:para}
\end{figure}

Our proposed method falls into the $1^{st}$ category and leverages the policies to learn a good machine translation model. We train the proposed model using data of different domains to search for parameters that can be easily adjusted to new domains. This is critical for domain-specific machine translation as the training data usually is limited.

Fig. \ref{fig:para} illustrates how to use meta learning to find a good initial parameter which can be easily adjusted to new domains. Methods like fine tuning firstly use an optimizer like SGD or Adam to learn a parameter $\theta$ which minimizes the loss function on a dataset of the general domain. And then starting from $\theta$, the model is further tuned to make the objective loss function reach a local optimum on the targeted domain and obtain the new parameter $\theta'$. The problem is, as stated above, the dataset of the target domain usually is too small for further tuning. If the starting point $\theta$ is far from the ``gold" parameter for domain specific NMT, we may end up with a $\theta'$ that is not fully optimized for MT on the target domain.
Using meta learning policy, we alternately optimize the model on different domains and eventually learn parameters which can be easily adjusted to new domains. 
Meta learning techniques have been adopted for multilingual machine translation \cite{gu2018meta}. This work adopts it for multiple domain machine translation.

\section{Method}
\begin{figure*}[ht]
  \centering
  \includegraphics[width=0.999\textwidth]{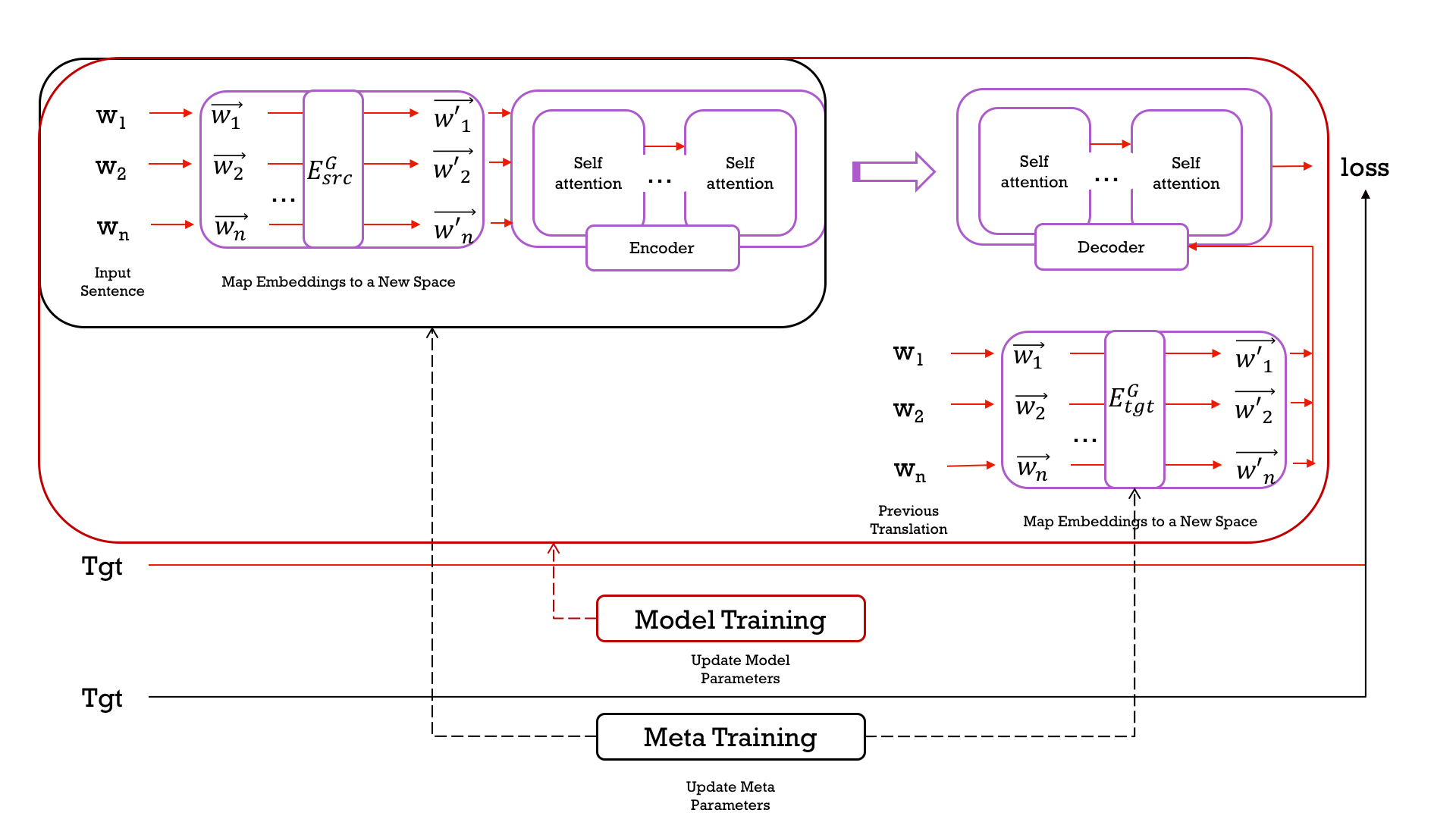}
  \caption{A graphical illustration of the proposed method. In training phase, source and target words are firstly projected to domain-invariant representational spaces and then are encoded/decoded. Parameters in the model are updated alternatively during training.}\label{fig:meta} 
\end{figure*}

The proposed method can be built in different NMT schemes. In this work we adopt the transformer model, a state-of-the-art NMT model for its performance and speed. Figure \ref{fig:meta} illustrates the architecture of the proposed model.
We first project words into a universal semantic space to reduce domain divergence in text representations. A new learning policy is used to update different parts of the NMT model to search for a parameter which can be easily adjusted to new domains. 

\subsection{Domain-Invariant Word Representation}
By training on the wiki crawl dataset with the fastText \footnote{https://fasttext.cc/docs/en/unsupervised-tutorial.html}, we obtain the word embeddings of the general domain, $E^G$, for source words. Note that the $E^G$ here is trained on general domain data and is not optimized for any of the specific domain we are working on nor is $E^G$ optimized for the machine translation task. Starting from $E^G$, we optimize it towards domain-specific machine translation. We keep the most frequent $n$ words in $E^G$ and use them as the basis in the word embedding space. Here $E^G$ is an $n*d$ matrix. The semantic space defined by $E^G$ is used as the domain-invariant semantic space and all words from different domains will be projected into this space.

For a word $w_k$, its representation in domain $i$ is written as $\Vec{w_k^i}$.
We map $\Vec{w_k^i}$ to the space defined by $E^G$ to obtain a new representation, $\Vec{w_k}$, for $w_k$. 

\begin{equation}
\begin{aligned}
        a_j^i&=\Vec{w_k^i} * A^i * E^G[j]\\
        \Vec{w_k}&=\sum_{j=1}^n a_j^i*E^G[j] \\
\end{aligned}
\end{equation}

$A^i$ is a $d*d$ matrix which is to be learned during the training. $\Vec{w_k}$ is new vector representation of $w_k$ and will be passed to the encoder. This projection helps reduce the divergence among inputs of different domains. It also makes it possible for identical tokens of different domains to have different representations. The same strategy is also adopted in the decoder.

\subsection{Encoding-Decoding}
The new word representations obtained through the above approach are further passed to the encoder. Though other seq2seq architectures could also be used, here we adopt the Transformer model \cite{vaswani2017attention} with multi-head self-attentions for its effectiveness.

At decoding, we also need to produce the domain-invariant embeddings for the target words. We adopt the same technique which is used for learning source word embeddings. The model first projects target word embeddings into a domain-invariant space in which all words are represented by a selected number of base word embeddings, and then passes the embeddings to the decoder.
 
\subsection{Learning Policy}

When migrated to a new domain, a good NMT model should be quickly adjusted using a limited amount of training data. To achieve this goal, we define two kinds of parameters, one kind being model parameters $\theta_0$ including all the parameters in the model, and the other kind being meta parameters $\theta_1$ which includes parameters in the transmission layers and the encoder. Meta parameters $\theta_1$ are tuned to reduce the domain divergence.
The following learning policy is adopted to update the two kinds of parameters.

For the training, we use datasets of several different domains $D^*$ of the same language pair. The data $D^i$ of each domain $i$ is split into three parts, a training set $D_{tr}^i$, a development set $D_{dev}^i$ and a test set $D_{te}^i$.

The learning procedure involves several training iterations with each iteration involving two parameter updates.
Firstly we sample a pair of domains $\{i, j\}$.
Then we use $D_{tr}^i, D_{dev}^i$ to update the model parameter $\theta_0$ in the model $M$. This step is referred to as model training. In the next step, using $D_{dev}^j$ we update only the meta parameter $\theta_1$ in $M$ and freeze the remains. As stated above, $\theta_1$ includes parameters in the transmission layers (of the source sides and target sides) and the encoder. As stated, meta parameters are used to fine tune the model for different domains. Intuitively it is sufficient to includes only transmission layers as meta parameters. Here we also include the encoder in the meta parameters $\theta_1$ as experiments and previous work \cite{li2016deep,gu2018meta} prove the usefulness of this strategy. Figure \ref{fig:para} illustrates how parameters are adjusted in this procedure. 

After several iterations, we obtain a series of new parameters $\theta_i'$. We collect all the gradients with respect to these parameters and use them to update the model as is done by \cite{finn2017model}.

\begin{algorithm}[ht]
\raggedright
\label{alg:A}
\begin{algorithmic}
\STATE {Datasets: $\{D^0,D^1,...,D^{n-1}\}$}

\STATE {Translation Model: $M(\theta)$}

\FOR{$D^i,D^j \in Datasets$}
\STATE $\theta_i \longleftarrow  \arg\min Loss(D_{tr}^i,D_{dev}^i; M(\theta))$ (Model training, update the translation model.)
\STATE $\theta_i' \longleftarrow \arg\min Loss(D_{dev}^j; M(\theta_i))$ (Meta training, update the transmission layers and the encoder.)
\ENDFOR
\STATE $\theta \longleftarrow \arg\min \sum_{i=0}^{n-1} Loss(D_{tr}^i; M(\theta_i'))$
\STATE $\theta \longleftarrow \arg\min Loss(D_{tr}^d,D_{dev}^d; M(\theta))$ (Fine-tune on $D_d$, the dataset of domain $d$.)
\RETURN $M(\theta)$

\end{algorithmic}
\caption{Learning Policy, Train MetaMT Using Data of Multiple Domains}
\end{algorithm}

Algorithm 1 describes the training procedure. Note that each training procedure involves two consecutive training processes. First, we conduct the model training using $D_{tr}^i, D_{dev}^i$ as the training and validation datset. Then we conduct the meta training with $D_{dev}^j$ split into two parts and used as the training set (90\%) and validation set (10\%). Note that although we use part of $D_{dev}^j$ as training data in meta training, it is essentially a redistribution of training data and no extra data has been exposed to $M$.

Each iteration mimics a domain adaptation of the proposed model. The meta parameter learns to handle domain divergence. Using meta learning, we obtain a model $M(\theta)$ which can be easily adjusted to match data of new domains. For NMT on a new domain $d$, we use $\theta$ as the initial parameters for $M$ and use $D^d$ to further train $M$ to get a domain specific model $M(\theta_d)$. Note that when fine-tuning onto the target domain, we update all the parameters $\theta$ in the model $M$.

\section{Experiments \& Analysis}

We conduct experiments to verify the effectiveness of the proposed model. We follow the same pre- and post- processing procedures for all the experiments unless otherwise stated.

\subsection{Data}
We use 7 En-Es parallel datasets of different domains for the evaluation. These datasets are all public available. For fair comparison, only subsets of these datasets with the same amount of sentence pairs are used to simulate NMT with limited data. However, this does not mean that the training data for our proposed model needs to be strictly balanced. The statistics of the datasets used in this work are shown in Table \ref{tab:data}.

\begin{table*}[ht]
\small
\centering
\begin{tabular}{|l|l|l|l|l|l|l|l|}\hline
 Data& EU Para & UN Data &GlobalVoices & JRCAcquis & EU Bookshops & OpenSub & Medline\\ \hline
 Sent. Pairs & 520K & 520K & 520K &520K &520K &520K & 143K\\ \hline
 En Tokens & 13.2M & 13.3M & 9.81M &10.11M &13.27M &3.34M &1.67M\\ \hline
 En Avg. Length & 25.41 & 25.58& 18.87& 19.45 &25.53  &6.44 &11.65\\ \hline
 Es Tokens & 13.69M & 15.82M & 10.58M & 11.36M&14.35M & 2.96M&1.89M\\ \hline
 Es Avg. Length &  26.43& 30.44 & 20.35 &21.86 &27.60 &5.71 &13.15\\ \hline
\end{tabular}
\caption{Statistics of En-Es Datasets Used in this Work} \label{tab:data}
\end{table*}

The JRCAcquis \cite{steinberger2006jrc} is a legal document dataset. Global Voices \cite{prokopidis2016parallel} is a collection of blogs on various topics. OpenSub is a dataset of Movie and TV subtitles \cite{lison2016opensubtitles2016}. The Europarll \cite{koehn2005europarl} and UN Para \cite{rafalovitch2009united} come from EU and UN proceedings. Medline \cite{liu2015translating} is a dataset constructed from biomedical articles from NIH’s MedlinePlus website.  EU Bookshops dataset \cite{skadicnvs2014billions} is a collection of publications in EU. All datasets are available on OPUS \cite{tiedemann2012parallel} \footnote{http://opus.nlpl.eu/}, the open parallel corpus, except the Medline dataset which is the biomedical domain $ESPAC_{MedlinePlus}$ corpus built in \cite{liu2015translating}.

\begin{table*}[ht]
\small
\centering
\begin{tabular}{|l|l|l|l|l|l|l|l|}\hline
 Data& EU Para & UN Data &GlobalVoices & JRCAcquis & EU Bookshops & OpenSub & Medline \\ \hline
 Transformer& 34.04 & 49.88 & 42.28 & 52.22 &20.78 &25.54 &50.95\\
 +Fine Tune & 37.10 & 54.81 & 45.24 & 60.07 &22.29 &27.31 &57.58 \\ \hline
 MetaMT     & 39.02 & 55.13 & 47.73 & 61.04 &22.74 &29.43 &59.06 \\ 
 -enc-proj  & 38.32 & 55.01 & 46.95 & 60.93 &22.25 &28.97 &58.77 \\ 
 -dec-proj  & 37.34 & 54.92 & 45.73 & 60.52 &21.02 &28.35 &58.22 \\ \hline
\end{tabular}
\caption{Performance Comparison (BLEU-4)} \label{tab:re}
\end{table*}

\subsection{Implementations}

Our proposed model is implemented using Pytorch \footnote{https://pytorch.org/}, a flexible framework for neural networks. We base our model on the transformer model \cite{vaswani2017attention} and the released Pytorch implementation \footnote{https://github.com/pytorch/fairseq}.
Parameters are set as follows:
word vector size = 300, hidden size = 512, number of layers=4, number of heads=6, dropout=0.3, batch size=64, and beam size=5. The pre-trained English and Spanish embedding are obtained using fastText \cite{mikolov2018advances} \footnote{https://fasttext.cc/} on the Wikipedia datasets of English and Spanish separately. We use the top 10K En/Es words as base words to construct the base semantic spaces.
At testing, we use beam search to find the best translated sentences. Decoding ends when every beam gives an $<EOS>$. 
\subsection{Baselines}
Various methods have been proposed for neural machine translation. Among them, we compare our methods against strong baselines.
\paragraph{Transformer}
We use Transformer as a strong baseline as it has achieved promising performances in several datasets.
We use a union of all the training data of different domain datasets for training.
\paragraph{Fine Tuning}
Fine-tuning is a practice used by transfer learning \cite{zoph2016transfer}. The model is firstly trained using available data and then fine-tuned using the task-specific dataset. As mentioned above, fine-tuning aims to find local minima for the loss function. It is stable and always achieves results comparable to other state-of-the-art systems \cite{chu2017empirical}.
\paragraph{MetaMT}
MetaMT is our proposed method. We also conduct ablation study by removing encoder side embedding projection and decoder side embedding projection, denoted as (-enc-proj,-dec-proj)

For the proposed model and the baselines, we use the same pre/post-processing and parameter settings for all methods in Table \ref{tab:re} and Table\ref{tab:re_ehr} unless otherwise stated.

We use byte pair encoding (BPE) to reduce the number of unknown words \cite{sennrich2015neural} for systems mentioned above (num of ops=20K).

\subsection{Results \& Analysis}
Our evaluation is done by a single reference, case-insensitive BLEU score using the Moses package \cite{papineni2002bleu}.
Results are reported in Table \ref{tab:re}.

As can be seen, the proposed MetaMT model yields gains in BLEU score of about 1.0 to 2.0 points, comparing with baselines except on a few datasets (UN Data and the EU Bookshop data). Both datasets cover a wide range of topics and contain many infrequent words, which may be one of the reasons that the improvement is not significant.

\subsection{Experiments on Very Small Dataset}

\begin{table*}[ht!]
\begin{center}
\small
\begin{tabular}{|l|c|c|c|c|c|}
\hline 
 & Sent. Pairs &  Word Tokens(EN) & Sent. Length (EN) &  Word Tokens(ES) & Sent. Length (ES)\\ \hline
Training & 2171 & 34534& 15.9& 36906&17.0 \\
Development &244 & 3946&16.2 & 4191& 17.2 \\
Testing & 595 &10201 &17.1 &10900 &18.3\\
\hline
\end{tabular}
\end{center}
\caption{\label{EHR_data} Statistics of EHR dataset.}
\end{table*}

\begin{table}[ht!]
\small
\centering
\begin{tabular}{|l|l|l|}\hline
Transformer & Transformer + Fine Tune & MetaMT \\ \hline
36.38 & 40.61 & 42.20\\ \hline
\end{tabular}
\caption{Performance Comparison (BLEU-4) on Electronic Health Record Dataset} \label{tab:re_ehr}
\end{table}

To further evaluate the performance of our proposed model, we also test our model on a very special dataset. 

We built an English-Spanish parallel electronic health record (EHR) notes corpus, which comprises 3,020 paired sentences from 57 de-identified EHR discharge summaries, randomly selected from patients with type 2 diabetes in a hospital. Translation was done by a professional medical translator whose first-language is Spanish, who spent over 1000 hours building the corpus, including back translation, a very costly task. Statistics of the EHR corpus are shown in Table \ref{EHR_data}. Results show that the proposed method, which is fine tuned by thousands of sentences is able to outperform transformer.

\section{Related Work}
NMT models require a large amount of parallel data for training their parameters from the very beginning \cite{bahdanau2014neural}. 
This becomes a severe problem for translation between low-resources languages or low-resource domains. Limited data results in a large quantity of words with low frequency which is hard to represent and translate. A lot of approaches have also been explored to learn the representations and translation of infrequent words. \cite{domhan2017using,DBLP:journals/corr/abs-1708-00712,sajjad2017neural,koehn2017six}.

Creating more data helps ease the infrequent word problem.
But according to the Zipf's law \cite{zipf2013psycho}, a few high token-frequency words account for most word occurrences in corpora. We often need a lot more data to increase the token frequencies of several infrequent words.

Besides, as previous work \cite{koehn2017six} showed, sometimes NMT models suffer on corpora with high proportions of frequent words. It seems NMT models favor words with moderate frequencies rather than those with extremely high or low frequencies.
Learning from mono-lingual data is another solution \cite{zhang2016exploiting,cheng2016semi,domhan2017using}. But much noise is introduced when using unsupervised learning. 

Meta learning \cite{DBLP:journals/corr/abs-1810-03548} explores the ability to learn automatically. It enables the model to learn from a limited number of data which makes it suitable for NMT with limited sources. If the model is facing a new task which overlaps with prior tasks, the model can quickly adjust itself to the new task according to its experiences. In NMT, researchers have tested a broad category of learning policies and obtained promising results \cite{dakwale2017fine,DBLP:journals/corr/MunkhdalaiY17,gu2018meta}.

\section{Conclusion}
We present a meta learning method for neural machine translation with limited resources. The proposed model uses a collection of base words to represent words from different domains in one semantic space to reduce domain divergence. Parameters in the model are defined into two groups and updated in a new learning policy. With the novel learning policy, data of different domains is used to update different parts of the model. Experiments verify the effectiveness of the proposed method in finding parameters which can be easily adjusted to new domains with only a limited number of training examples. 

The proposed training policy is not limited to neural machine translation as it does not rely on any model specific features or strategies.
In the future we will further investigate applying the meta training policy to other neural machine learning models.

\section{Acknowledgement}
This work was supported by the grant R01DA045816, R01HL125089, R01HL137794, R01HL135219, and R01LM012817 from the National Institutes of Health (NIH). The contents of this paper do not represent the views of the NIH.
\bibliographystyle{aaai}
\bibliography{mybib}

\end{document}